\documentclass[10pt,twocolumn,letterpaper]{article}

\usepackage{cvpr}              % To produce the CAMERA-READY version
\usepackage[dvipsnames]{xcolor}

\definecolor{cvprblue}{rgb}{0.21,0.49,0.74}
\definecolor{yd}{RGB}{40,40,200}
\usepackage[pagebackref,breaklinks,colorlinks,citecolor=cvprblue]{hyperref}

\title{Enhanced Visual Instruction Tuning with \\ Synthesized Image-Dialogue Data }

\def\SP{~~}

\author{
Yanda Li$^{1,2 \ast}$,
\SP
Chi Zhang$^2$\thanks{Equal contributions. Work was done when Yanda Li was a Research Intern at Tencent.}~\thanks{Project Leader }, 
\SP 
Gang Yu$^2$\thanks{Corresponding Author \\Project page: \url{https://github.com/icoz69/StableLLAVA}},
\SP
Zhibin Wang$^2$,
\SP 
Bin Fu$^2$, \\
\SP
Guosheng Lin$^{3}$,
\SP
Chunhua Shen$^{4}$,
\SP
Ling Chen$^{1}$,
\SP
Yunchao Wei$^{5 \ddagger}$
\\[0.1125cm]
\normalsize 
$ ^1$ University of Technology Sydney
\SP ~~~
$ ^2 $ Tencent
\SP ~~~
$ ^3 $ Nanyang Technological University\\
\normalsize 
\SP ~~~
$ ^4 $ Zhejiang University
\SP ~~~
$ ^5$ Beijing Jiaotong University \\
\\
{\normalsize e-mail: $^1$\normalsize \tt   liyanda95@gmail.com, $^1$ling.chen@uts.edu.au }\\
$^2$\normalsize \tt  \{johnczhang, skicyyu, billzbwang, brianfu\}@tencent.com \\
$^3$\normalsize \tt  gslin@ntu.edu.sg
$^4$\normalsize \tt  chunhua@me.com
$^5$\normalsize \tt  wychao1987@gmail.com
}

\begin{document}
\makeatletter
\let\@oldmaketitle\@maketitle% Store \@maketitle
\renewcommand{\@maketitle}{\@oldmaketitle% Update \@maketitle to insert...
 \centering
    \includegraphics[width=\textwidth]{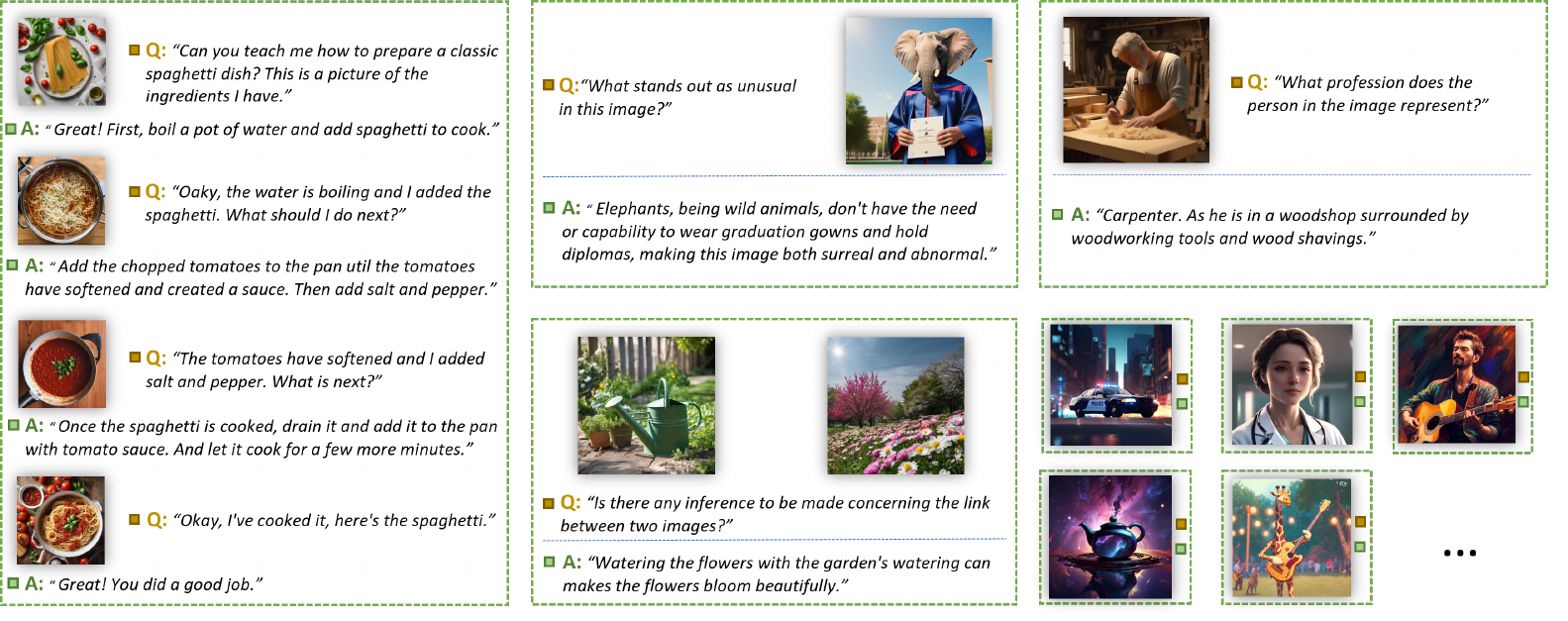}
  %  \vspace{-1 em}
     \captionof{figure}{\textbf{
     %Demo
     Examples of synthesized visual instruction data}.
     We use ChatGPT and text-to-image generation models to synthesize various forms of visual instruction tuning data, such as multi-round dialogue data, multi-image reasoning data, and anomaly detection data. These data are used to train the multimodal large language models.
}
    \label{fig:teaser}
    %\vspace{-1 em}
    \bigskip}                   % ... an image
\makeatother
\maketitle

\begin{abstract}

The remarkable multimodal capabilities demonstrated by OpenAI's GPT-4 have sparked significant interest in the development of multimodal Large Language Models (LLMs). A primary research objective of such models is to  align visual and textual modalities effectively while comprehending human instructions.
 Current methodologies often rely on annotations derived from benchmark datasets to construct image-dialogue datasets for training purposes, akin to instruction tuning in LLMs. However,  these datasets often exhibit domain bias, potentially constraining the generative capabilities of the models. In an effort to mitigate these limitations, we propose a novel data collection methodology that synchronously synthesizes images and dialogues for visual instruction tuning. This approach harnesses the power of generative models, marrying the abilities of ChatGPT and text-to-image generative models to yield a diverse and controllable dataset with varied image content. Additionally, datasets can be arbitrarily scaled. This not only provides greater flexibility compared to existing methodologies but also significantly enhances several model capabilities. Our research includes comprehensive experiments conducted on various datasets. The results emphasize substantial enhancements in more than ten commonly assessed capabilities. Additionally, our model achieves state-of-the-art results across multiple widely recognized multimodal benchmarks.

\end{abstract}

\section{Introduction}

The advent of OpenAI's ChatGPT\cite{openai_chatgpt} sets a significant advancement in the realm of Artificial Intelligence (AI), revealing a range of impressive abilities embedded in Large Language Models (LLMs).
These models, exemplified by GPT-4\cite{openai2023gpt}, showcase exceptional versatility by handling not just images but also excelling in tasks once thought impossible.
This includes understanding humor within images and drafting website code from basic sketches, aspects that highlight its revolutionary potential.

Despite these remarkable achievements, a critical aspect remains undisclosed: the specific mechanics underpinning GPT-4, particularly concerning the seamless integration of multimodal information into LLMs. This knowledge gap has spurred a concerted research effort to address this conundrum. 

Notably, an emerging method receiving considerable attention involves the utilization of adapter-based techniques~\cite{zhang2023llama,gao2023llama,luo2023cheap}, which allow the training of a visual-to-text adapter that can convert features from pre-trained visual models into LLM tokens. Intriguingly, these adapter-based methods have demonstrated efficacy comparable to the undisclosed GPT-4 model.
The appeal of adapter-based methodologies lies in their utilization of the extensive knowledge embedded in pre-trained large visual models and LLMs. A key focus is on training a lightweight adapter, thereby sidestepping the computationally intensive process of training visual models and LLMs from scratch. This targeted resource allocation enhances efficiency by concentrating computational efforts on areas yielding the most substantial outcomes.

A prerequisite for implementing these frameworks is the availability of paired vision-text image data. This dual-purpose dataset aligns visual and textual features, fostering an understanding of human instructions crucial for generating appropriate responses. Analogous to instruction tuning in LLMs~\cite{wei2021finetuned}, this process is commonly referred to as visual instruction tuning. Existing methods~\cite{dai2023instructblip,liu2023visual,peng2023instruction,liu2023aligning,chung2022scaling} typically construct visual instruction tuning datasets by leveraging established vision datasets, extracting information such as image captions, spatial locations, and categories to form dialogues. This approach maximizes resource utilization, creating a comprehensive and efficient training dataset for multimodal LLMs.

However, despite the efficiency and simplicity of this approach to dataset construction,  certain limitations still persist. Existing large-scale vision-text datasets, such as LAION~\cite{schuhmann2022laion} and CC~\cite{changpinyo2021conceptual}, often contain noise. Consequently, training only a subset may inadequately align visual-text features for immediate user requirements. 
Moreover, benchmark datasets~\cite{changpinyo2021conceptual,schuhmann2022laion, lin2014microsoft} often exhibit a domain bias, primarily in terms of image styles. For instance, prevalent datasets such as COCO~\cite{lin2014microsoft} predominantly feature images from everyday life, while stylized images like cartoons are rarely represented. Additional, these vision annotations may also constrain the types of dialogues generated from them. For example, none of the current datasets contain data to directly enhance the model's ability to comprehend jokes presented in the images, an impressive feature of GPT-4. 
Besides, as multi-image dialogues become increasingly integral to the practical application of multimodal LLMs, the current lack of comprehensive multi-image datasets further underscores the need for data enrichment in this domain.

\begin{figure*}[t]
\centering
\includegraphics[width=1.0\textwidth]{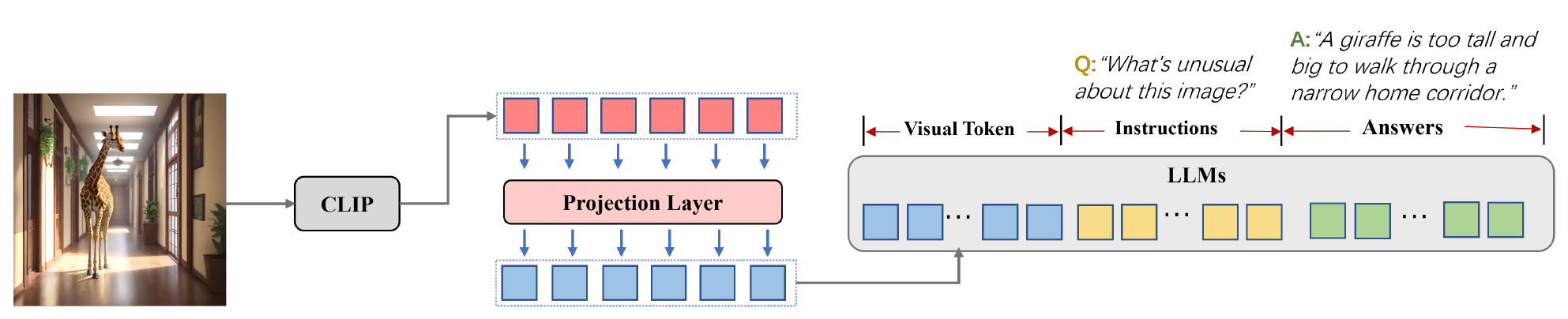}
% \vspace{-0.5 em}
\caption{\textbf{Architecture of LLaVA}. We use the open-source LLaVA model as a testbed for our proposed data generation pipeline. The model is trained to predict the next tokens in the answers given the visual tokens and instruction tokens in an auto-regressive manner.  } 
\label{fig:llava}
% \vspace{-1.5em}
\end{figure*}

In addressing these limitations, we propose a novel data collection approach to enhance visual instruction tuning. Building upon recent successes in the AI-Generated Content (AIGC) field, we leverage generative models to produce image-dialogue pair data for visual instruction tuning.
More concretely, we employ ChatGPT to create data that includes image-generating prompts and content-based dialogues. We then utilize the text-to-image diffusion model, StableDiffusion~\cite{rombach2022high}, to generate images based on these prompts. Finally,  the synthesized images and generated dialogues are employed to train multimodal LLMs.
%By generating both the image content and dialogues simultaneously, we can produce diverse training data and exercise greater control over its nature and quality. 
Simultaneously generating both image content and dialogues enables the production of diverse training data, affording greater control over its nature and quality. \emph{This flexibility allows us to construct multi-turn dialogues and datasets involving multi-image reasoning, which are challenging to obtain from other benchmarks.} 
% Moreover, due to the flexibility of our approach, we can construct multi-turn  dialogues and data involving multi-image reasoning, which are challenging to obtain from other benchmarks.

Additionally, our methodology can potentially integrate more advanced image generative models, 
such as DALLE3~\cite{BetkerGohJingBrooks:DALL-E3}, 
to provide higher-level control over image contents like specifying complex spacial relations.
This advanced control could generate more complex instructions to enhance image understanding capabilities. Examples from our synthesized visual instruction tuning datasets are shown in Figure~\ref{fig:teaser}. Building upon the flexible pipeline outlined above, users can tailor the generation of data to enhance specific capabilities based on their task requirements. 
Furthermore, our method of generating both images and dialogues eliminates constraints on data volume, thereby facilitating the production for limitless scaling of the datasets.

To demonstrate the effectiveness of our proposed pipeline, we conducted extensive experiments. Our main contributions are as threefold:
\begin{itemize}
\item We develop a novel pipeline for generating visual instruction tuning datasets by leveraging text-to-image diffusion models.
\item To showcase its flexibility, we have built a dataset with various form of capabilities including multi-image data, and our results have shown improvements across all abilities.
\item Extensive experimental analysis on multiple benchmarks shows the effectiveness of the proposed method, outperforming baseline and existing SOTA approaches.

\end{itemize}

\section{Related Work}
Recent research~\cite{zhu2023minigpt,liu2023visual,ye2023mplug} efforts in multimodal Large Language Models (LLMs) have yielded promising strategies to efficiently align the embeddings of other modalities with language tokens. This has made it possible to effectively utilize pre-trained encoders from other modalities and LLMs, which effectively reduces the computational burden and training time. While there are alternative research approaches that include training-free  methods leverage expert models~\cite{wu2023visual,shen2023hugginggpt,yang2023gpt4tools}, these are not the focus of our work here.

\textbf{Adapter-based LLMs.}
One of the promising areas in multimodal LLMs research is the development of adapter-based methods\cite{zhang2023llama,gao2023llama,luo2023cheap,liu2023visual,zhu2023minigpt,ye2023mplug,yang2023exploring}. These methods aim to bridge different modalities with a learnable interface requiring minimal training efforts. The advantage of such methods is the ability to use pre-trained encoders from other modalities and LLMs as initial parameters, minimizing the necessity of training from scratch.
Design variations in these adapters also exist. Some studies, for instance, have directly trained project layers to align embeddings to language tokens. As an example, the LLaVA model~\cite{liu2023visual} learns a simple linear layer to translate the image features to language tokens, a strategy that has yielded surprisingly good results. In contrast, other works~\cite{ye2023mplug} have employed learnable queries to extract information from the embeddings of other modalities, an approach first used in the Flamingo model~\cite{alayrac2022flamingo}. Further diversifying the field, the LLaMA-Adapter~\cite{zhang2023llama} introduces a lightweight adapter module for parameter-efficient tuning, and LaVIN~\cite{luo2023cheap} designs a mixture-of-modality adapter to process instructions of different modalities.

\textbf{Visual instruction tuning datasets.}
Preparing multimodal training datasets, such as image-text pairs, plays a crucial role in enabling the training of multimodal LLMs. The main objectives are to align the modalities and to enable the model to follow human instructions. Most existing methods~\cite{liu2023visual,rotstein2023fusecap,li2023mimic,yin2023lamm,li2023m,liu2023aligning,zhang2023llavar} construct a visual instruction tuning dataset based on benchmark datasets, leveraging the wealth of information provided by the annotations.
The LLaVA model~\cite{liu2023visual}, for example, uses GPT4 to construct datasets based on the annotations of the COCO dataset\cite{lin2014microsoft}. This approach utilizes image captions and bounding box coordinates to generate various forms of dialogues, such as descriptions of image content and inferences of spatial relations. However, these constructed datasets are restricted by the information contained in the annotations, such as a fixed set of categories, which may limit their diversity.

In contrast, our approach takes advantage of image generation models that have been well-trained on large-scale datasets. These models provide image data that is controllable in various aspects, allowing us to enhance particular capabilities of multimodal LLMs with our pipelines. Furthermore, our framework can be extended by integrating more advanced generative models that support various forms of guidance tailored to specific needs.

\begin{figure*}[t]
\centering
\includegraphics[width=0.8\textwidth]{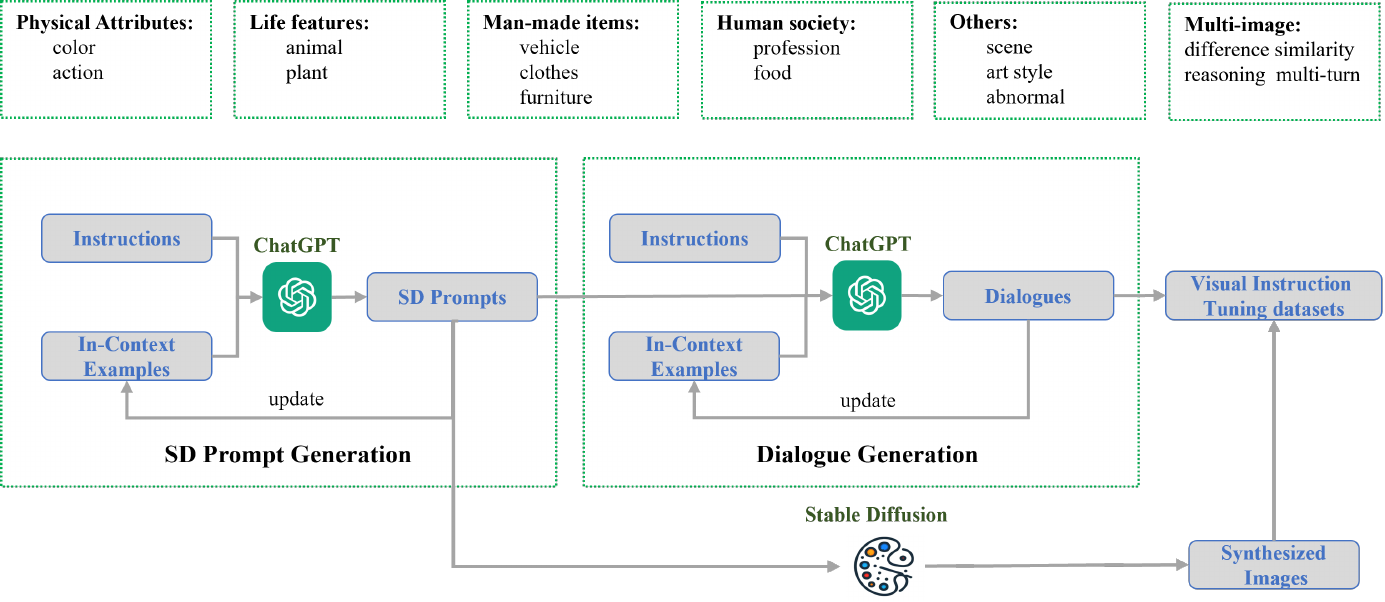}
% \vspace{-0.5 em}
\caption{\textbf{Our proposed pipeline for generating visual instruction tuning datasets. } We instruct ChatGPT to generate both StableDiffusion prompts and the associated dialogues.  For specific generation templates, please refer to the supplementary materials. } 
\label{fig:pipeline}
\vspace{-3mm}
% \vspace{-1.5em}
\end{figure*}

\section{Preliminary}
\label{sec:pre}
In order to gauge the effectiveness of our data generation strategy, we have elected to utilize the open-sourced LLaVA~\cite{liu2023visual,liu2023improved} model as our multimodal LLM model. It should be noted that our pipeline is model-agnostic, meaning the datasets generated via our approach can be employed for training a variety of other models. To set the stage for a detailed exposition of our pipeline, this preliminary section provides an overview of the structure and training strategies used in the LLaVA model. For a more in-depth understanding, the reader may refer to the original publication~\cite{liu2023visual}.

%model replacement

\textbf{Architecture.}
The architecture of the LLaVA model elegantly combines a pre-trained LLM, specifically the Vicuna-13B~\cite{chiang2023vicuna}, and a pre-trained visual encoder, known as CLIP-ViT-L/14~\cite{radford2021learning}. This fusion of text and visual processing abilities is facilitated by the incorporation of a learnable linear layer. This linear layer has the primary task of projecting the image features, developed by the CLIP encoder, into the word embedding space of the LLM.  The resultant projected embeddings effectively function as tokens within the LLM, creating a synergy between text and visual data streams. It's worth mentioning that in LLaVA-1.5~\cite{liu2023improved}, the linear projection layer has been substituted with a two-layer MLP. Additionally, the LLM model has been replaced with Vicuna-1.5-13B, and the input image size has been increased from 224x224 to 336x336, thereby elevating the model's multimodal capabilities to a greater extent. A detailed illustration of this model structure can be found in Figure~\ref{fig:llava} of this paper.

\textbf{Training and datasets.}
The core of LLaVA's training process revolves around visual instruction tuning, which necessitates a triplet of data inputs: images, questions, and corresponding answers. The goal here is to prompt the model to predict the next tokens in the answers conditioned on visual tokens and instruction tokens (\eg, a question) in an auto-regressive manner. 

The training of LLaVA is split into two distinct stages. Each stage has a specific focus and uses different data and optimized parameters. The first stage is primed for modality alignment and utilizes a dataset of 595K image-text pairs. In this stage, the training is concentrated on optimizing the parameters of the linear layer, with the weights of the visual encoder and LLM kept constant, ensuring the focus remains on aligning visual and text modalities.
In the second stage, the optimization process is expanded to also include the LLM's weights. This stage of training employs a smaller set of 158K multimodal dialogue data, derived from the COCO datasets. By fine-tuning the parameters of both the linear layer and the LLM, the model can effectively align and utilize both visual and text modalities, setting the stage for visual instruction tuning.

Building on the foundation established by LLaVA, LLaVA-1.5 has significantly augmented the dataset by incorporating additional data from sources like Region-level VQA~\cite{kazemzadeh2014referitgame,krishna2017visual,mao2016generation} and GQA~\cite{hudson2019gqa}, resulting in the expansion of the second-stage data to 665K examples.

By understanding the LLaVA model's architecture and training process, we can better understand and appreciate the implications of our proposed data generation methodology.

\begin{figure*}[t]
\centering
\includegraphics[width=\textwidth]{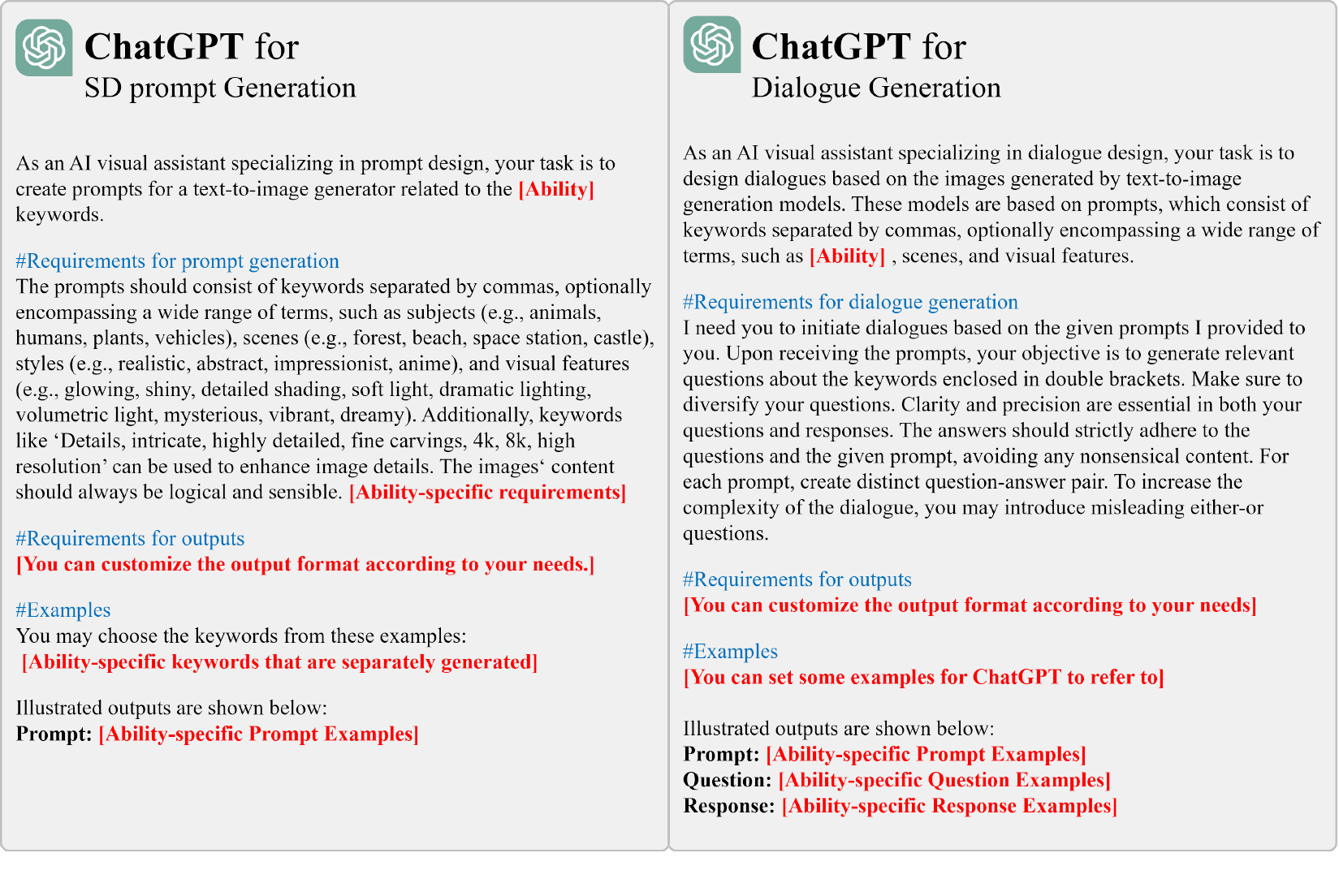}
% \vspace{-0.5 em}
\caption{\textbf{Templates for guiding ChatGPT to generate StableDiffusion prompts (left) and dialogues (right).  } Content in \textcolor{red}{\textbf{red}} represents ability-specific information. We only provide an example template for constructing dialogues regarding a single image in this figure. For additional forms of data, such as multi-image reasoning and multi-turn dialogues, please refer to our supplementary materials. } 
\label{fig:templete}
\vspace{-3mm}
% \vspace{-1.5em}
\end{figure*}

\section{Methods}
In this section, we describe our pipeline for generating visual instruction tuning datasets. Our methodology stands distinct from previous data collection methods in its dual-generation approach, synthesizing not only 
the images but also the associated dialogues. The schematic representation of our proposed methodology is depicted in Figure.~\ref{fig:pipeline}.
The following subsections provide an in-depth discussion of each component.

\subsection{Image Generation}
Generating images with StableDiffusion~\cite{rombach2022high} relies on the use of prompts, typically comprising several weighted keywords, with those placed at the beginning given higher precedence during image generation.
These keywords cover various image aspects such as the subject, scene, style, and other visual elements, \eg, image quality and lighting. 
Additional emphasis can be added through the use of brackets.
To encourage diversity and stability during image generation, we add capability-specific instructions and cautions during prompting ChatGPT.
For instance, in the task of generating images for joke understanding, we direct ChatGPT to create prompts that would result in the generation of abnormal images, like a ``\emph{giraffe walking through a narrow corridor}'', which are unlikely to be found in reality.
When generating multi-image data, pairs of prompts can be generated concurrently based on predefined specific criteria.
%\zhangchi{similarly... for multi images...}
For maximum effect, we ensure that the most crucial keywords are placed at the beginning of the generated prompts, which are double-bracketed for additional emphasis. 
Furthermore, we instruct ChatGPT to avoid generating prompts that are non-visual, such as the act of growing. 
%\zhangchi{left an example of instruction template for prompt generation} 
The instruction template for prompt generation is provided in the left part of Figure~
\ref{fig:templete}.
The generated prompts are then utilized with StableDiffusion to produce visually realistic images, which are subsequently encoded by LLaVA's vision encoder into visual tokens for LLMs.

\subsection{Dialogue Generation}
Following the generation of images, we employ ChatGPT to generate dialogues based on the same prompts used for image synthesis.
As outlined in Section~\ref{sec:pre}, LLaVA's dual training stages have distinct objectives: the initial stage seeks alignment between visual and textual data, while the subsequent stage focuses on understanding diverse human instructions. 
Accordingly, we design dialogues that cater to both stages. For the first stage, dialogues center on image descriptions based on image generation prompts.  Here, we provide a set of pre-defined image description questions separately generated by ChatGPT, and only instruct ChatGPT to generate the answers. Particularly, for enhancing joke understanding, we instruct ChatGPT to describe what is funny or abnormal in the image. 
Taking the example of the ``\emph{giraffe walking through a narrow corridor}'', a representative dialogue might be:  ``\emph{Question: What is unusual in the image? Response: In reality, a giraffe is too tall and big to walk through a narrow home corridor.}'' The detailed instruction template for dialogue generation is shown in the right part of Figure~\ref{fig:templete}.
%\zhangchi{The detailed instruction template for dialogue generation is shown in supplementary materials.}

\begin{figure*}[t]
\centering
\includegraphics[width=0.8\textwidth]{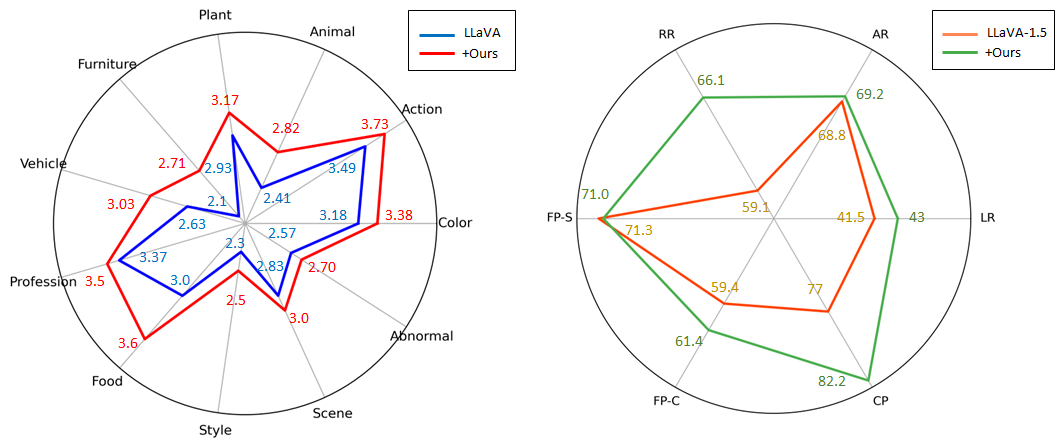}
% \vspace{-0.5 em}
\caption{ Left: Results on evaluation benchmarks for various abilities. Right: Comparison of various subcategories on MMBench~\cite{liu2023mmbench} with the baseline. Our model outperforms the baselines on both benchmark datasets.
} 
\label{fig:real}
\vspace{-2mm}
% \vspace{-1.5em}
\end{figure*}

%\zhangchi{
For the second stage, we mainly focus on generating data for enhancing multi-image reasoning abilities, which can be difficult to derive from existing benchmark datasets.
Our multi-image ability focuses on the reasoning of similarity, difference, and logical relations. 
Additionally, we construct image-text interleaved multi-turn dialogues. In
this process, ChatGPT is prompted to concurrently generate image prompts and dialogues, detailed template can be found in the supplementary materials.
%with each prompt being constructed by extracting keywords derived from the ongoing dialogue text.
% we construct multi-image multi-turn interleaved dialogues... }.   
we encourage ChatGPT to generate a diverse set of question types, including yes-or-no questions, wh-questions, and either-or questions. Moreover, we caution ChatGPT against generating inherently ambiguous questions and encourage the generation of clear, accurate, and unambiguous responses.

\subsection{In-Context Examples}
It's been observed that ChatGPT possesses in-context learning capabilities, meaning it can understand and grasp the essence of a task given a few examples. To leverage this, we
add in-context examples during the generation of both StableDiffusion prompts and dialogues. 

During the data generation process, we observed that ChatGPT sometimes produced a lack of diversity. For example, when generating colors, the outputs frequently revolved around common color categories. To overcome this, we independently generate ability-related keywords such as color categories with ChatGPT, and utilize them as a reference during the prompting process. This additional step promotes a more diverse range of prompts, thereby enriching our visual instruction tuning dataset.

In a bid to actualize this diversity, we have adopted a dynamic approach throughout the generation process. Specifically, at regular intervals, we implement a substitution strategy where a certain fraction of the original in-context examples are replaced with data that has been newly generated. This strategy ensures that ChatGPT, during its continuous generation process, not only maintains but progressively amplifies the diversity within the dataset. Concurrently, it significantly mitigates the risk of over-repetition of certain categories and styles. This balanced approach aids in the creation of a more comprehensive and representative dataset.

\section{Experiments}
In this section, we detail the experiments conducted to validate the effectiveness of our novel data collection approach for visual instruction tuning. We will first introduce the datasets we used for training, followed by the evaluation datasets, evaluation strategy, and finally present our quantitative and qualitative results.

\subsection{Training Datasets}
In order to fully illustrate the flexibility and comprehensive applicability of our proposed pipeline, we proceeded to generate a diverse and expansive dataset covering various form of commonly assessed single-image abilities. These capabilities span a wide spectrum from foundational image recognition tasks to more nuanced aspects of visual reasoning. Specifically, our selected abilities include recognition and understanding of physical attributes, life features, man-made items, human society and others for training in the first stage.

Each ability's dataset was formulated following a standard template, illustrated in supplementary material. Our synthesized dataset includes approximately 126K image-dialogue pairs. 
In addition, we also generated a dataset of 3K multi-image instances, encompassing descriptions of image similarity, difference, logical relations, and multi-turn dialogue data for the second stage.
These datasets, in combination with the raw LLaVA dataset, provides a comprehensive training set in our experiments.

\subsection{Evaluation Metrics}
\textbf{Evaluation datasets.}
To demonstrate our performance more clearly, we tested on a series of public multi-modal datasets, including VisWiz~\cite{gurari2018vizwiz}, MM-Vet~\cite{yu2023mm}, MME~\cite{fu2023mme}, and MMBench~\cite{liu2023mmbench}.

Subsequently, we constructed a multi-image test set consisting of 30 dialogues to assess the models' performance on this specific data type. This dataset evaluates the models across differences, similarities, and reasoning relationships among the images. The test data was sourced from publicly available datasets and manually annotated.

Simultaneously, we devised a real-image benchmark to assess the efficacy of our training process across various capabilities. These benchmarks comprehensively cover the full spectrum of the diverse single-image abilities, serving as a robust evaluation foundation. This benchmark comprises real images sourced manually from publicly accessible repositories. For each of abilities, we meticulously curated 30 testing images, each paired with a manually annotated question-answer dialogue, resulting in a total of 330 test samples.

\textbf{Evaluation Strategy}
In terms of the evaluation process, we employ different testing strategies depending on the benchmarks used. 

For publicly accessible multimodal test sets, such as VizWiz~\cite{gurari2018vizwiz}, MMBench~\cite{liu2023mmbench}, we adhere to official guidelines by downloading the designated test data. Subsequently, we convert the data into a model-compatible format, store the model outputs, and adapt them into the standardized testing format. Following this, we employ official testing scripts or submit the results to the official testing website for assessment, presenting the ultimate results. These publicly available datasets employ accuracy as the evaluation metric. For specific evaluation rules, please refer to their respective papers for details.

In evaluating the diverse capabilities and multi-image benchmarks we've generated, we leverage GPT-4~\cite{openai2023gpt} to assist in scoring model outputs. The specific instruction template used for result evaluation is illustrated in supplementary material. We have established six scoring levels, ranging from 0 to 5. Each score level is accompanied by detailed descriptions of the evaluation criteria, and we assist GPT-4 in better assessment by providing a series of scoring examples. 
In particular, a score of 0 indicates that the predicted answer has no relevance to the reference answer, while a score of 5 signifies that the predicted answer aligns seamlessly with the annotated reference answer without any deviation. Drawing on our manual annotations, we conduct evaluations on the results produced by each model. The average GPT score within each test set serves as the ultimate metric for our benchmark evaluations.

\subsection{Quantitative comparison to state-of-the-arts}

\textbf{Public multimodal benchmarks}
We perform quantitative performance comparisons against various state-of-the-art methods on different benchmarks, as illustrated in Table~\ref{tab:comp}. Utilizing LLaVA-1.5-13B as the baseline, we integrate our synthesized data with its original dataset for training. Training is carried out with identical parameter configurations as LLaVA-1.5. The outcomes demonstrate substantial improvements on many benchmarks, emphasizing the enhanced performance achieved by our approach.
zh

\begin{table}[t]
\caption{Quantitative comparison with other state-of-the-arts methods on multiple multimodal benchmarks. We achieve state-of-the-art performance on four benchmarks.}
\centering
\resizebox{0.9\columnwidth}{!}{
\begin{tabular}{l|cccc}
\toprule
Method &  
  VisWiz & MM-Vet & MME & MMB \\
  \midrule
  BLIP2~\cite{li2023blip} & 19.6 & 22.4 & 1293.8 & - \\
  InstructBLIP~\cite{dai2023instructblip} & 33.4 & 25.6 & 1212.8 & -  \\
  IDEFICS-9B~\cite{idefics} & 35.5 & - & - & 48.2 \\
  IDEFICS-80B & 36.0 & - & - & 54.5 \\
  Qwen-VL~\cite{bai2023qwen} & 35.2 & - & - & 38.2 \\
  Qwen-VL-Chat & 38.9 & - & 1487.5 & 60.6 \\
  LLaVA-1.5~\cite{liu2023improved} & 53.6 & 35.4 & 1531.3 & 67.7 \\
  \textbf{Ours} & \textbf{58.4} & \textbf{36.1} & \textbf{1532.3} & \textbf{69.1} \\ 
 \bottomrule
\end{tabular}
\label{tab:comp}
}

% \vspace{-12mm}
\end{table}

\textbf{Multi-image benchmark}
To validate the effectiveness of multi-image capabilities, we manually curated a benchmark of real images. This test set assessed performance based on differences, similarities, and logical relations. The evaluation metric used was the GPT-4 score mentioned. 
We used LLaVA-13B as the baseline and incorporated multi-image data in the second training phase. Since LLaVA itself lacks the capability for multi-image input, we modified the testing code for LLaVA to enable it to accept multiple sets of images simultaneously. The comparison with LLaVA results is shown in the Table~\ref{tab:multi}, indicating a notable improvement across various multi-image capabilities despite adding less multi-image data in the process.

%We compared our results with the baseline LLaVA, as shown in Table~\ref{tab:multi}. Our model achieved the best results on the multi-image benchmark.

\begin{figure*}[t]
\centering
\includegraphics[width=\textwidth]{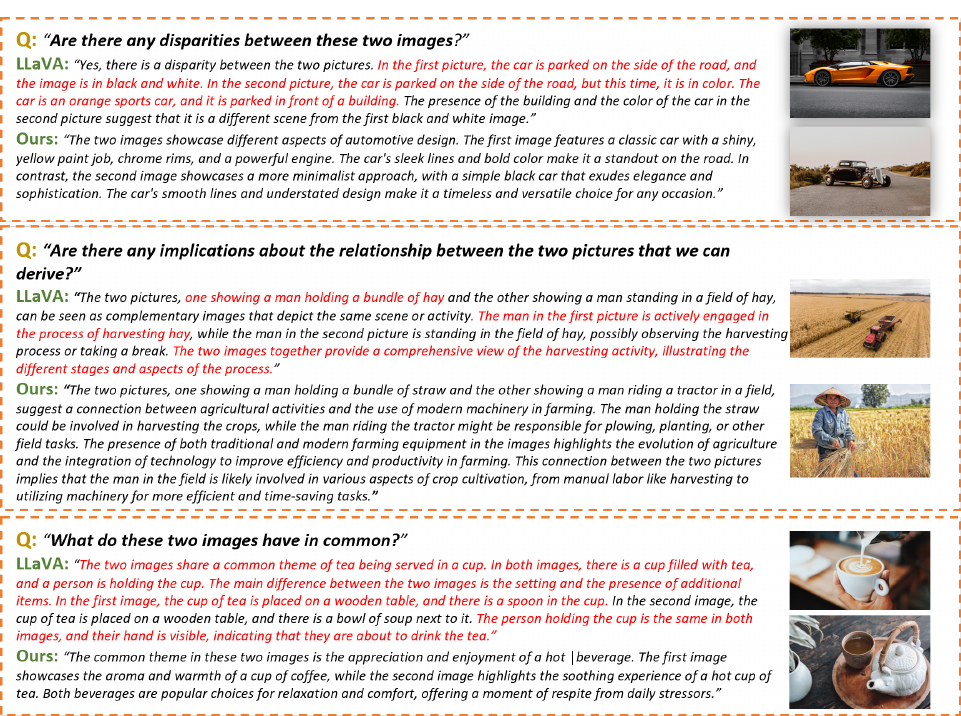}
% \vspace{-0.5 em}
\caption{\textbf{ Comparison of the results generated by LLaVA and our trained model. }  Content in \textcolor{red}{\textbf{red}} represents inaccurate information. Our model can better adhere to question instructions, rendering more precise answers.
} 
\label{fig:comp}
\vspace{-3mm}
\end{figure*}

\textbf{Comparison of various abilities.}
To validate the effectiveness of our generated data, we conducted comprehensive tests on distinct capabilities using a meticulously designed testing benchmark. Employing LLaVA-13B as our baseline, the quantitative comparison of the baseline results and ours is shown in the left part of Figure~\ref{fig:real}. 
Notably, our trained model consistently outperforms the LLaVA-13B baseline across all various capabilities on real-image benchmarks, which suggests the synthesized datasets' generalizability and our pipeline's robustness.

\begin{table}[t]
\caption{Quantitative comparison on the multi-image benchmark. After the addition of multi-image data, various multi-image capabilities have significantly improved.}
\centering
\resizebox{\columnwidth}{!}{
\begin{tabular}{l|cccc}
\toprule
Method &  Difference & Similarity & Logical relations & Average \\
  \midrule
  % Flamingo & & & & \\
  % Otter & & & &  \\
  LLaVA~\cite{liu2023visual} & 2.7 & 2.2 & 3.1 & 2.67\\
  Ours & \textbf{3.6} & \textbf{2.8} & \textbf{3.7} & \textbf{3.37} \\
 \bottomrule
\end{tabular}
\label{tab:multi}
}
\vspace{-3mm}
% \vspace{-12mm}
\end{table}

To better validate our superiority, we conduct a comparison of subcategory performance on MMBench, using the LLaVA-1.5-13B as the baseline. 
The tested subcategories in MMBench encompass six aspects: attribute reasoning (AR), coarse perception (CP), fine-grained perception (cross-instance) (FP-C), fine-grained perception (instance-level) (FP-S), logic reasoning (LR), and relation reasoning (RR).
The final results are shown in the right part of Figure~\ref{fig:real}, indicating better performance of subcategory on MMBench, which also attests to the high quality of our generated data.

\subsection{Qualitative results}

Supplementing the quantitative analysis, we provide a qualitative comparison between our model's results and LLaVA-13B in Figure.~\ref{fig:comp} on multi-image data. Our model exhibits a heightened ability to adhere to question instructions, rendering more precise answers. 

We compare our approach with the LLaVA-13B baseline, revealing its limitations: it struggles to differentiate between multi-image contents and provides incomplete answers to questions. Our method, incorporating multi-image data, enhances the model's understanding of multiple images, demonstrating its effectiveness. 
Additional qualitative results will be included in the supplementary materials.

\section{Conclusion and Future Work}

In the rapidly evolving realm of Large Language Models, efficiently integrating multimodal information is a key research focus. In this study, we introduced an innovative data collection method to enhance visual instruction tuning for multimodal models. Compared to existing strategies, our approach uniquely combines image and dialogue generation, effectively addressing limitations found in benchmark datasets. By carefully crafting instruction templates, our method ensures high-quality training data covering a broad range of crucial capabilities for multimodal models.

Our research opens avenues for exploration. Moving forward, we aim to leverage advanced generative models to enhance model abilities, including spatial comprehension and fine-grained recognition. With promising results from our dual-generation method, forward-thinking data collection techniques are poised to play a significant role in the future of LLM research.

\textbf{Limitations} Due to constraints in text-to-image models like stable diffusion, generating certain data types, such as text-rich images and tables, is not effective in the current pipeline. We anticipate these constraints will be addressed with ongoing advancements in text-to-image generation techniques.

{
    \small
    \bibliographystyle{ieeenat_fullname}
    \bibliography{main}
}

\newpage
\setcounter{section}{0}

\section{Training Details}
During the model training phase, we employed the original LLaVA~\cite{liu2023visual} configuration as the foundation for our training process. In both stages, we utilized 8 NVIDIA V100 GPUs. To conserve GPU memory, we employed deepspeed with zero3 during model training, disabling tf32 and opting for fp16. The remaining parameters, including epochs and learning rates, were set according to the original LLaVA configuration. For specific parameter details, please refer to the the original publication~\cite{liu2023visual}.

\section{Data Filtering Mechanism}
To enhance the quality and diversity of generated data, we implemented a data filtering mechanism. The filtering process involved the following aspects:

\begin{enumerate}
    \item We filtered the generated data based on the repetition rate of extracted prompts. The filtered prompts were then used to generate corresponding images and dialogues.
    
    \item Filtering was also conducted based on prompt and dialogue length. We observed that limiting the number of prompt keywords to 10 or fewer was essential. Excessively long prompts could hinder the text-to-image model's ability to accurately identify key generation points, potentially leading to the generation of erroneous and chaotic images. 
    For generated dialogues, we imposed length restrictions of 500 characters to prevent  disorderly outputs by ChatGPT.
    
    \item For certain categories, we implemented restrictions based on specific attributes of the capabilities. For example, when generating content related to construction workers, the model tended to focus on buildings. To address this, additional human attributes were incorporated into the prompts to ensure a more accurate representation.
\end{enumerate}
% \begin{table}[t]
% \caption{Quantitative comparison on the multi-image benchmark. After the addition of multi-image data, various multi-image capabilities have significantly improved.}
% \centering
% \resizebox{\columnwidth}{!}{
% \begin{tabular}{l|cccc}
% \toprule
% Method &  Difference & Similarity & Logical relations & Average \\
%   \midrule
%   % Flamingo & & & & \\
%   % Otter & & & &  \\
%   LLaVA~\cite{liu2023visual} & 2.7 & 2.2 & 3.1 & 2.67\\
%   Ours & \textbf{3.6} & \textbf{2.8} & \textbf{3.7} & \textbf{3.37} \\
%  \bottomrule
% \end{tabular}
% \label{tab:multi}
% }

% \vspace{-3mm}
% % \vspace{-12mm}
% \end{table}
\section{Multi-image Generation Template}
\begin{figure*}[t]
\centering
\includegraphics[width=0.7\textwidth]{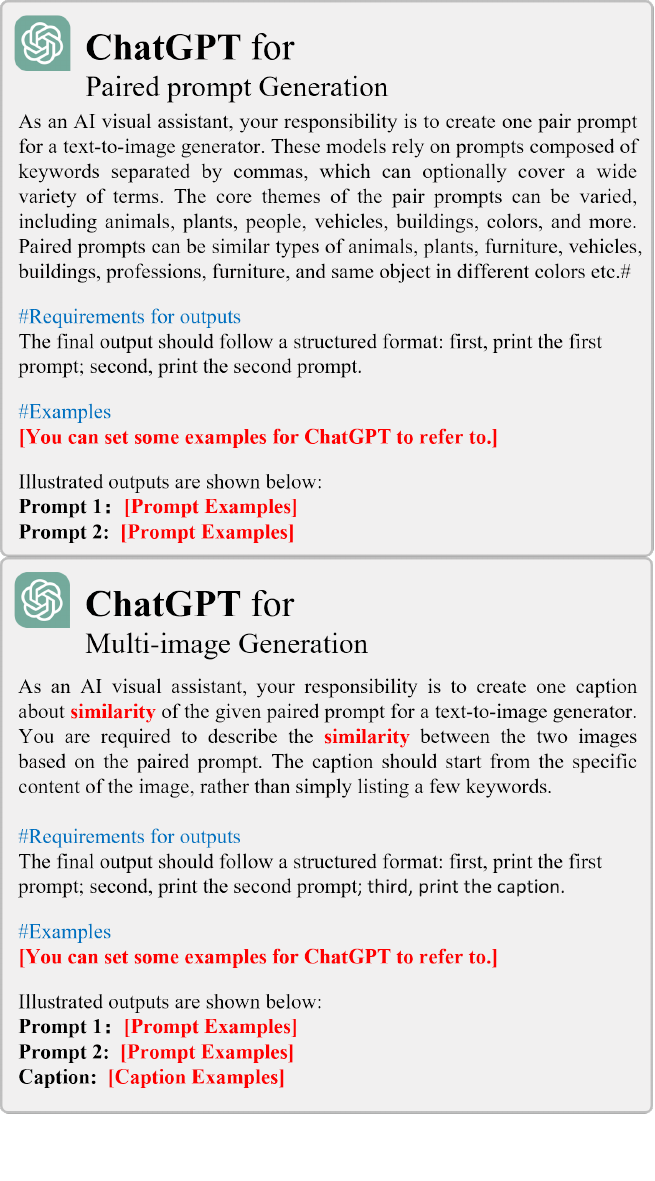}
\caption{Instruction template for multi-image data generation. } 
\label{fig:multi-image}
\end{figure*}

We have illustrated the instruction template for generating multi-image data in Figure~\ref{fig:multi-image}. To enhance the diversity of the data, we first use ChatGPT to generate a series of paired prompts. Subsequently, based on these paired prompts, we generate corresponding captions. As shown in the Figure~\ref{fig:multi-image}, we can generate different types of data by configuring different examples and replacing similarities with difference or logical relations.

\section{Interleaved multi-turn dialogues Generation Template}
\begin{figure*}[t]
\centering
\includegraphics[width=0.7\textwidth]{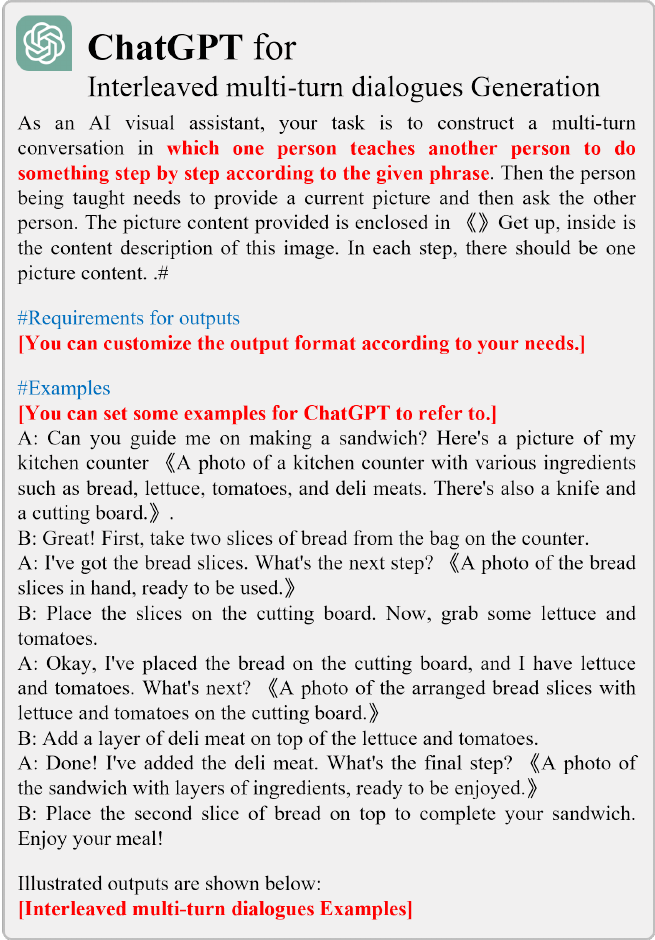}
\caption{Instruction template for interleave multi-turn dialogues generation. } 
\label{fig:inter}
\end{figure*}

We present a template for generating interleaved multi-turn dialogues as Figure~\ref{fig:inter}. Similar to multi-image generation, we randomly generate a series of non-repeating phases to enhance data diversity. Based on these phases, we generate interleaved multi-turn dialogues. The red sections are replaceable. In our generated dialogues, the focus is primarily on instructional content related to the use of everyday objects and recipes. Users can make specific adjustments according to their needs. Finally, the brackets part represents the prompt used for stable diffusion to generate images. During the model training process, the content within the brackets will be replaced with an \textbf{image\_placeholder}.

\section{GPT-4 Scoring Criteria}
\begin{figure*}[t]
\centering
\includegraphics[width=0.9\textwidth]{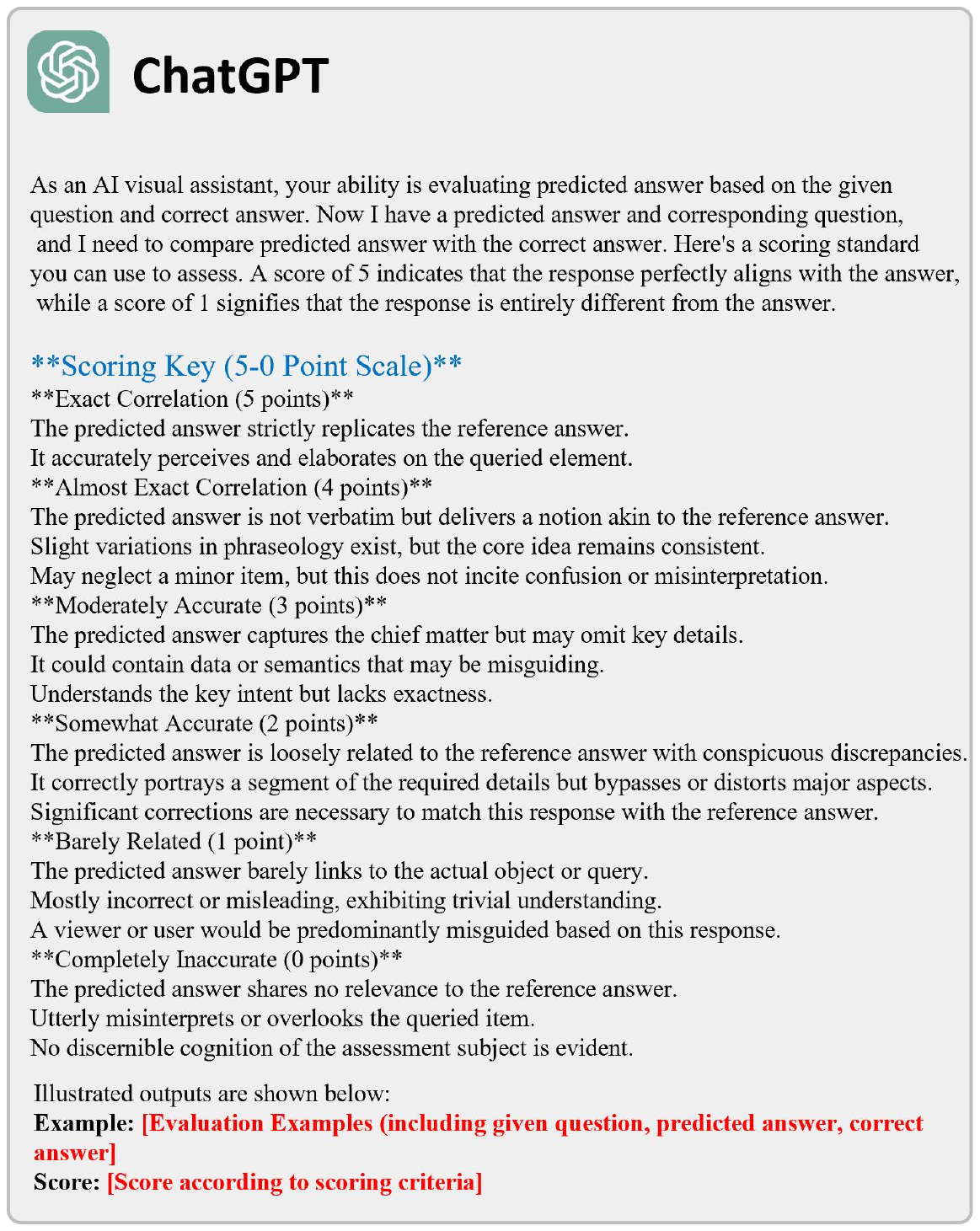}
\caption{Score criteria based on GPT-4. } 
\label{fig:score}
\end{figure*}

As shown in the Figure~\ref{fig:score}, we present our detailed GPT-4 scoring system. We have established a scale of 0-5 with six levels of scores, and for each score, we provide detailed evaluation criteria along with specific examples for assessment. Utilizing the template in the Figure~\ref{fig:score}, evaluations are conducted for each model, and the average of the results is taken as the final score.

\section{More qualitative Results.}

\begin{figure*}[t]
\centering
\includegraphics[width=\textwidth]{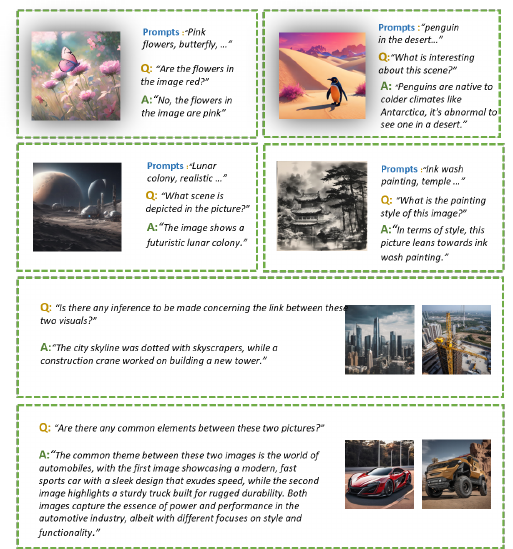}
% \vspace{-0.5 em}
\caption{\textbf{Qualitative results:}
Demonstrations of diverse data we generated. } 
\label{fig:supp_exam}
% \vspace{-1.5em}
\end{figure*}

\begin{figure*}[t]
\centering
\includegraphics[width=\textwidth]{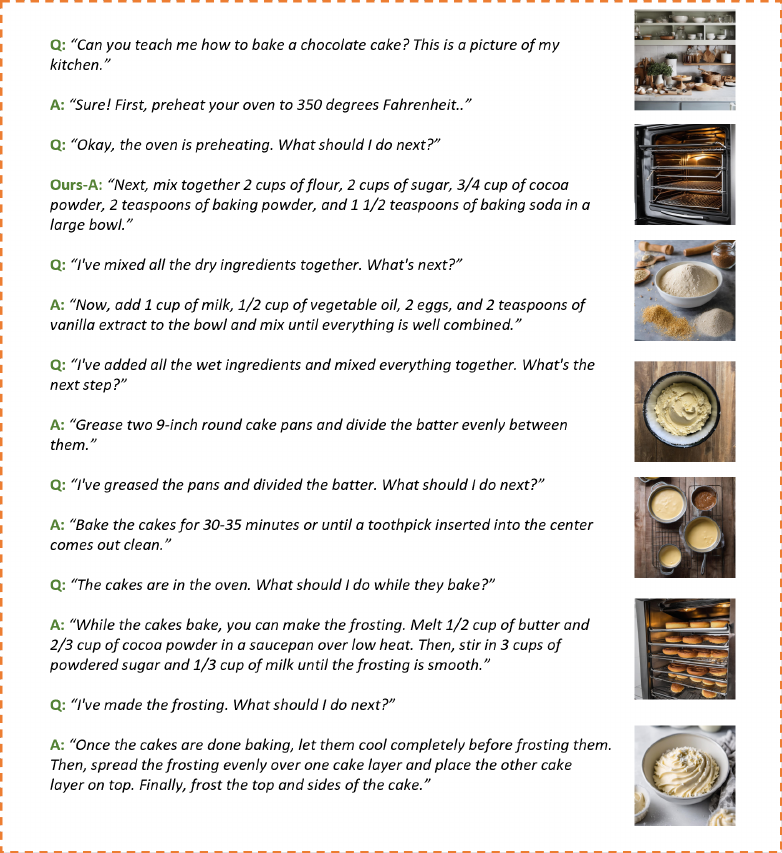}
% \vspace{-0.5 em}
\caption{\textbf{Qualitative results:}
Demonstrations of diverse data we generated. } 
\label{fig:supp_exam2}
% \vspace{-1.5em}
\end{figure*}

We showcase more generated data as shown in the Figure~\ref{fig:supp_exam}~\ref{fig:supp_exam2}, which includes various capabilities, as well as multi-image dialogues and interleaved multi-turn dialogues, to validate the effectiveness and diversity of our pipeline.

\begin{figure*}[t]
\centering
\includegraphics[width=0.9\textwidth]{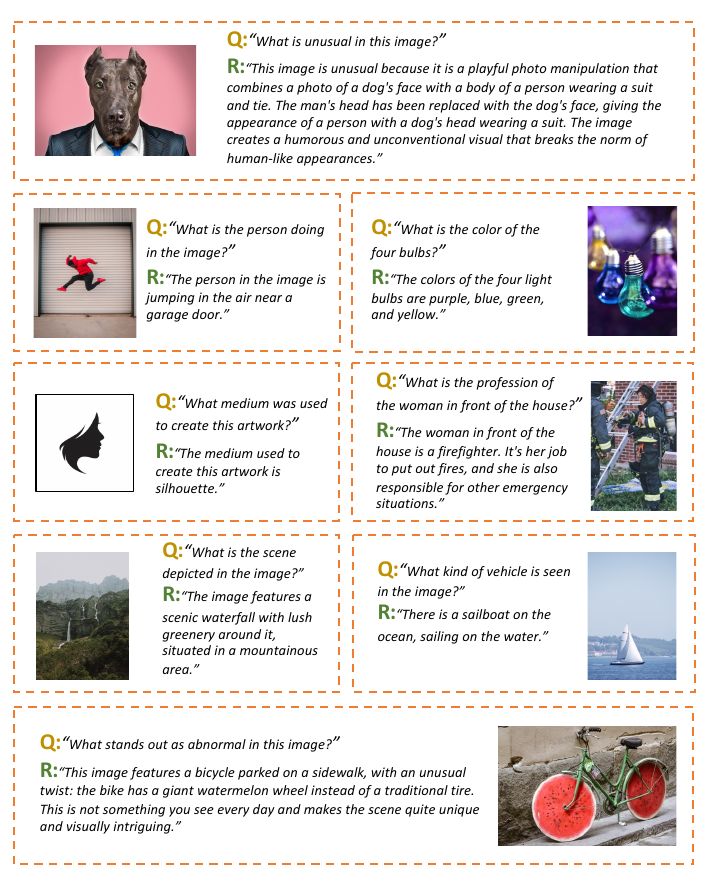}
% \vspace{-0.5 em}
\caption{\textbf{Qualitative results:}
Demonstrations of our method's effectiveness across diverse real-world image scenarios. } 
\label{fig:case}
% \vspace{-1.5em}
\end{figure*}

We present additional outcomes of our methodology on more real images in Figure~\ref{fig:case}. Spanning diverse capabilities like profession recognition, color categorization, or anomaly detection, our approach consistently exhibits efficacy.
In particular, our method demonstrates a robust capability in discerning abnormal images.

\begin{figure*}[t]
\centering
\includegraphics[width=0.9\textwidth]{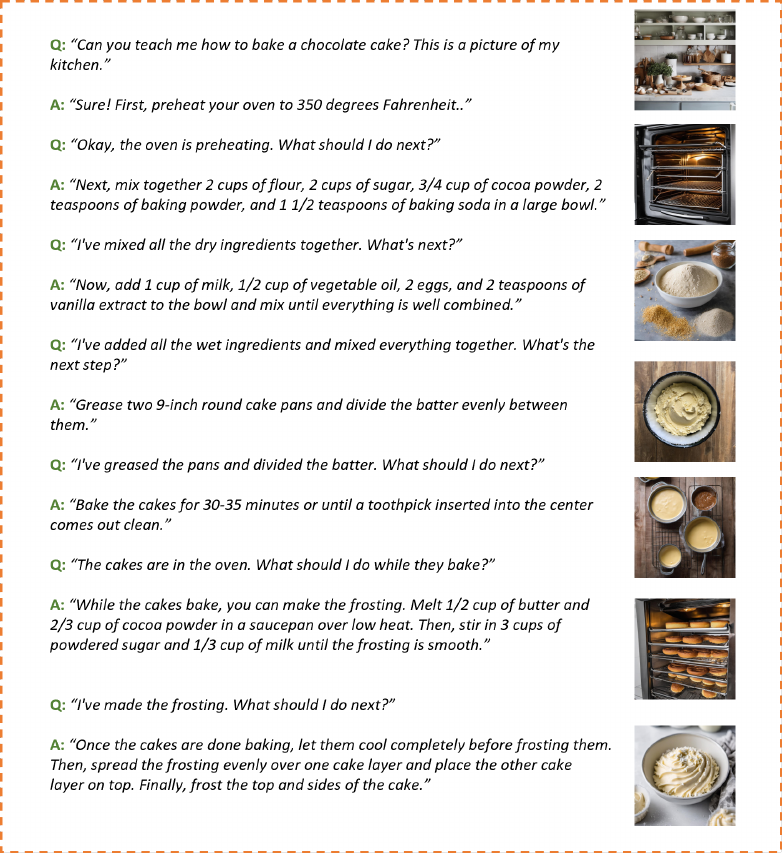}
% \vspace{-0.5 em}
\caption{\textbf{Qualitative results:}
Demonstrations of our method's effectiveness across interleaved multi-turn dialogues. } 
\label{fig:dia}
% \vspace{-1.5em}
\end{figure*}

\begin{figure*}[t]
\centering
\includegraphics[width=0.9\textwidth]{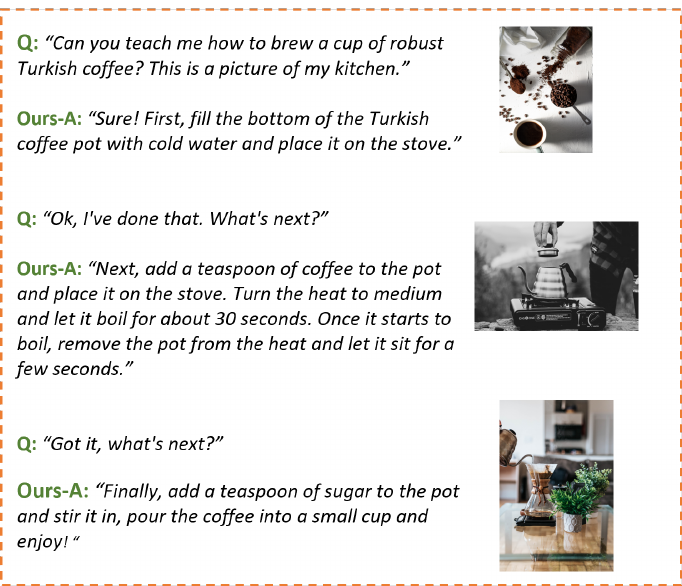}
% \vspace{-0.5 em}
\caption{\textbf{Qualitative results:}
Demonstrations of our method's effectiveness across interleaved multi-turn dialogues. } 
\label{fig:dia2}
% \vspace{-1.5em}
\end{figure*}

Simultaneously, we validated our model on multi-turn interactive dialogues, and the results are shown in the Figure~\ref{fig:dia}~\ref{fig:dia2}. It can be observed that our model can provide detailed descriptions of each step, which is attributed to the high-quality data we generated.

\end{document}